\documentclass[preprint,11pt]{elsarticle}

\usepackage{amssymb}
\usepackage{amsmath}
\usepackage{array,multirow}
\newcolumntype{P}[1]{>{\centering\arraybackslash}p{#1}}
\usepackage{caption}
\usepackage{graphicx}
\usepackage{rotating}
\usepackage{subfig}
\usepackage{lineno}
\usepackage{xcolor}
\usepackage{environ}
\usepackage{changepage}
\usepackage{color}
\usepackage[boxruled,commentsnumbered]{algorithm2e}
\NewEnviron{myequation}{%
\begin{equation}
\scalebox{0.8}{$\BODY$}
\end{equation}
}

\date{}

\begin{document}

\begin{frontmatter}

\title{Efficient Learning of Pinball TWSVM using Privileged Information and its applications}
\author[mymainaddress]{Reshma Rastogi (nee. Khemchandani)\corref{mycorrespondingauthor}}
\cortext[mycorrespondingauthor]{Corresponding author}
\ead{reshma.khemchandani@sau.ac.in}

\author[mymainaddress]{Aman Pal}
\address[mymainaddress]{Department of Computer Science,\\
South Asian University\\
New Delhi, India\\}

\date{}
\begin{abstract}
In any learning framework, an expert knowledge always plays a crucial role. But, in the field of machine learning, the knowledge offered by an expert is rarely used. Moreover, machine learning algorithms (SVM based) generally use hinge loss function which is sensitive towards the noise.  Thus, in order to get the advantage from an expert knowledge and to reduce the sensitivity towards the noise, in this paper, we propose privileged information based Twin Pinball Support Vector Machine classifier (Pin-TWSVMPI) where expert's knowledge is in the form of privileged information. The proposed Pin-TWSVMPI incorporates privileged information by using correcting function so as to obtain two nonparallel decision hyperplanes. Further, in order to make computations more efficient and fast, we use Sequential Minimal Optimization (SMO) technique for obtaining the classifier and have also shown its application for Pedestrian detection and Handwritten digit recognition. Further, for UCI datasets, we first implement a procedure which extracts privileged information from the features of the dataset which are then further utilized by Pin-TWSVMPI that leads to enhancement in classification accuracy with comparatively lesser computational time. 
\end{abstract}


\end{frontmatter}


\section{Introduction}

Classification is a well-studied problem in machine learning. However, learning still remains a challenging task especially when the amount of training data is limited \cite{lapin2014learning}. Hence, an expert's knowledge in addition to the training data can play a significant role in achieving the overall performance improvement. In the classical supervised learning paradigm, the learner's goal is to find a decision function, obtained using given training data along with a small generalization error on the unknown test data. In practice, knowledge offered by an expert is rarely used. However, a lot of expert's knowledge including explanations, comments, comparisons, etc. along with training data are also available in human learning. In \cite{vapnik2009new}, Vapnik et al. termed this additional prior information as the privileged information, which is available only at the training stage but not available while testing. Vapnik et al., \cite{vapnik2009new} proposed a Learning Using Privileged Information(LUPI) also known as Support Vector Machine with Privilege Information(SVMPI), which can boost the convergence rate of learning especially when the learning problem itself is NP-hard \cite{vapnik2009new}.

In the last few decades, Support Vector Machine (SVM) has become a powerful tool for classification problem \cite{xie2018uncertain,phan2018automatically,saha2013combining}. Motivated by the statistical learning theory and the maximum margin principle, SVM \cite{cortes1995support,vapnik1998statistical} tries to find an optimal separating hyperplane by solving a quadratic programming problem (QPP). However, solving QPP is expensive especially for large-scale datasets. To improve the performance of SVM, in \cite{khemchandani2007twin}, Jayadeva et al. proposed twin SVM (TWSVM) for binary classification in the spirit of the Generalized Eigenvalue Proximal Support Vector Machine (GEPSVM) \cite{fung2005multicategory}. TWSVM aims to identify two nonparallel hyperplanes such that each hyperplane is proximal to its own class and away from the samples of other class and its obtain by solving a pair of smaller-sized QPPs. The training speed of the TWSVM is faster than the standard SVM because it solves two smaller sized QPPs rather than a single larger-sized QPP. Twin support vector machines have gained importance in recent past and have been extended to several work like \cite{bala,nasiri,Mei,gupta}.

Recently, Huang et al. \cite{huang2014support}, proposed SVM with the pinball loss function (Pin-SVM) where hinge loss function was replaced with pinball loss function, which brought noise insensitivity to the classifier \cite{huang2014support}. Huang et al., \cite{huang2014support} have also shown that for noise-corrupted data, Pin-SVM is theoretically and empirically better than SVM. Recently, in \cite{mehrkanoon2014non}, Mehrkanoon et al. have presented non-parallel support vector classifiers with different loss functions. They have also shown the superiority of pinball loss function over hinge loss function in TWSVM.

In modern research, enhancing machines with the ability to interact with the human have become one of the most interesting and potentially useful challenges. Machine learning plays a key role in order to design tool to detect and track people \cite{nasiri2014energy}. In recent years, pedestrian detection has become a popular topic of research due to its several applications \cite{zhang2016far}. Real world applications \cite{min2018recognition} in pedestrian detection include robotics, surveillance, content-based indexing and automatic driver-assistance systems in vehicles. In this paper, we perform the experiment on pedestrian detection to demonstrate the effectiveness of privileged information with the proposed formulation. It is essential to note that in computer vision research problems, time plays a crucial role as it usually involves nearly real-time interaction with the environment which leads to a large number of samples. Thus, our proposed formulation shows a significant deduction concerning training time while maintaining the overall performance of the method.

In the past few years, extracting privileged/expert information from the benchmark datasets has become a challenging task. Recently, in \cite{qi2014new}, Qi et al. have proposed a fast version of TWSVM based privileged information termed as FTWSVMPI where the authors present the privileged information as the solution of two additional QPP model along with two more QPP of TWSVM. This leads to major disadvantage of Qi et al. \cite{qi2014new} model as it increases the training time of the classifier. To overcome this drawback of Qi et al. model, in our previous work \cite{pal2018learning}, we have proposed a novel method to extract privileged information from the datasets itself which avoid solving two additional QPPs. Further, Qi et al. \cite{qi2014new} and our previous model \cite{pal2018learning} have used hinge loss function which leads to noise sensitivity. Therefore, in this paper, we propose a novel twin support-vector machine with the pinball loss function based on privileged information, which brings back the noise sensitivity to the classifier. The key features of this paper are listed below:

\begin{itemize}
\item  We propose Pin-TWSVM with privileged information by replacing error variable with privileged information variable in the optimization problem of Pin-TWSVM \cite{mehrkanoon2014non}. The privileged information variable is obtained by using correction space which defines correcting function \cite{vapnik2009new}. 

\item Firstly, in order to take care of structural risk component, a regularization term in the objective function is introduced.

\item The dual problems of Pin-TWSVMPI have the similar formulation with that of standard SVM, and hence they are solved efficiently by using SMO \cite{platt1998sequential} which makes computation faster as compared to FTWSVMPI.

\item We also utilize a novel method proposed by Pal and Rastogi \cite{pal2018learning}, which extract privileged information from the features of datasets where expert's knowledge is not available, e.g., in UCI datasets.

\item Experimental results on several benchmark UCI datasets \cite{asuncion2007uci} indicate that the proposed formulation achieves better classification accuracy as compared to other formulation discuss in this paper and with considerably lesser computational time for both linear as well as nonlinear cases.

\item To establish the efficacy of the propose formulation with privileged information, we have shown its application for pedestrian detection over INRIA dataset \cite{dalal2005histograms} and Handwritten digit recognition over MNIST dataset.
\end{itemize}

This rest of the paper is organized as follows. Section 2 discusses the Pin-TWSVM. The Pin-TWSVMPI is proposed in section 3, which includes both the linear and nonlinear cases. Section 4 discusses the privileged information used for both the applications such as pedestrian detection and handwritten digit recognition. Experiments on UCI benchmark, INRIA and MNIST datasets are conducted to verify the effectiveness of propose Pin-TWSVMPI in section 5. Section 5.1 discusses the method used to extract privileged information from the dataset. Section 6 provides the concluding remarks.

\section{Related Works}
We consider a binary classification problem with a training set $T=\{(x_1,y_1),(x_2,y_2),...,(x_l,y_l)\}$ and privileged information $T^*=\{(x^*_1,y_1),(x^*_2,y_2\\),...,(x^*_l,y_l)\}$, where $X_l=\{x_i:i=1,2,....,l\}$ and $X^*_l=\{x^*_i:i=1,2,....,l\}$  are the \textit{l} data points in \textit{d} and $\textit{d}^*$ dimensions with corresponding class labels $Y_l=\{y_i\in[1,-1] :i=1,2,...,l\}$ respectively.  
  
 Data points belonging to class $+1$ and $-1$ are represented by matrices \textit{A} and \textit{B} each with number of patterns $m_1$ and $m_2$, respectively. Therefore, the size of matrices \textit{A} and \textit{B} are $(m_1\times d)$ and $(m_2\times d)$, respectively. Here $d$ is the dimension of the feature space. Let  $\textit{A}_i$$(i=1,2,...,m_1)$ and $\textit{B}_i$$(i=1,2,...,m_2)$ are the row vectors in \textit{d}-dimensional real space $\mathbb{R}^d$ that represents feature vector of class $+1$ and class $-1$ data samples respectively. 
 
 Privilege information belonging to class $+1$ and $-1$ are represented by matrices $\textit{A}^*$ and $\textit{B}^*$ each with number of patterns $m^*_1$ and $m^*_2$, respectively. Therefore, the size of matrices $\textit{A}^*$ and $\textit{B}^*$ are $(m^*_1\times d^*)$ and $(m^*_2\times d^*)$, respectively. Here $d^*$ is the dimension of privileged information feature space. Let  $\textit{A}^*_i$$(i=1,2,...,m^*_1)$ and $\textit{B}^*_i$$(i=1,2,...,m^*_2)$ are the row vectors in $\textit{d}^*$-dimensional correcting space $\mathbb{R}^{d^*}$ that represents privileged information of class $+1$ and class $-1$ respectively.
%

 The privileged information paradigm can be described as follows: given a set of triplets (training data) \vspace{0.2cm} \\ \vspace{0.2cm}  
 $(x_1,x_1^{*},y_l),...,(x_l,x_l^{*},y_l), ~~~~~x_i\in X_l, ~x_i^*\in X^*_l, ~y_i\in \{-1,1\} $\\
 where $X^*$ is the correcting space which is obtained by expert knowledge. The classical paradigm of supervised machine learning is described as follows: given a set of pairs \vspace{0.2cm} \\
 $(x_1,y_1),...,(x_l,y_l),~~~~~x_i\in X,~y_i\in \{-1,1\}$.\\
 
 In real-world, an expert always supply privilege information $x^*$ in the correcting space of $X^*$ with the admissible set of the correcting functions \cite{vapnik2009new}.

 \subsection{Twin Support Vector Machine with Pinball Loss function}
\begin{figure}[htp]
                     \centering
                     \subfloat[Linear Loss]{\includegraphics[width = 1.5in,height=1.3in]{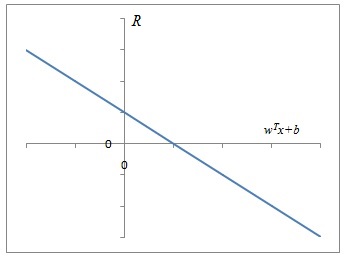}}~~~~
                     \subfloat[Hinge Loss]{\includegraphics[width = 1.5in,height=1.3in]{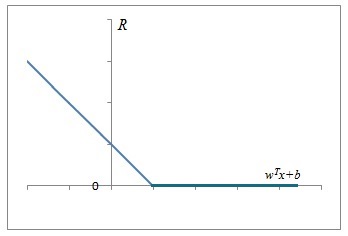}}~~~~\\
                     \subfloat[Quadratic Loss ]{\includegraphics[width = 1.5in,height=1.3in]{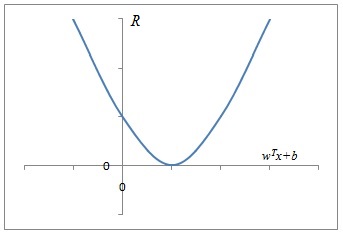}}~~~~
                     \subfloat[Pinball Loss]{\includegraphics[width = 1.5in,height=1.3in]{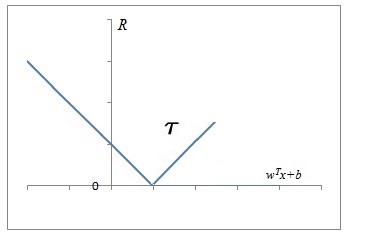}} 
                     \caption{Geometrical interpretation of different loss functions } 
                     \label{Fig:Loss_functions}
               \end{figure}

 In \cite{huang2014support}, Huang et al. proposed a SVM classifier with pinball loss function, named as Pin-SVM. With the help of pinball loss function, the authors use the quantile distance to measure the margin and propose the corresponding classifier to maximize the quantile distance.  
  The pinball loss function is given as follows:
  \begin{eqnarray}
  L_\tau(x,y,f(x))=
  \begin{cases}
      1-yf(x) & :  if ~~1-yf(x)\geq 0,\\
      -\tau(1-yf(x)) & : if ~~ 1-yf(x)<0,
    \end{cases}
  \end{eqnarray}
  which can be regarded as generalized $L_1$-loss function. Figure \ref{Fig:Loss_functions} shows the geometric interpretation of different types of loss functions. Figure \ref{Fig:Loss_functions}(d) represents pinball loss function.  
               
  Taking motivation from \cite{khemchandani2007twin} and \cite{huang2014support}, in \cite{mehrkanoon2014non}, Mehrkanoon et al. have proposed TWSVM with pinball loss function termed as Pin-TWSVM. The corresponding model is formulated as the following pair of QPPs:
 \begin{eqnarray}
 \label{eq:Pin-TWSVM1}
 \underset{w_1,\;b_1,\;\xi_2}{Min} & \frac{1}{2}||Aw_1+e_1b_1||^2_2+c_1e_2^T\xi_2\nonumber\\
 s.t. & -(Bw_1+e_2b_1)\geq e_2-\xi_2,\\
 & -(Bw_1+e_2b_1)\leq e_2+\frac{1}{\tau}\xi_2,\nonumber
 \end{eqnarray} 
 and
 \begin{eqnarray}
  \label{eq:Pin-TWSVM2}
  \underset{w_2,\;b_2,\;\xi_1}{Min} & \frac{1}{2}||Bw_2+e_2b_2||^2_2+c_2e_1^T\xi_1\nonumber\\
  s.t. & (Aw_2+e_1b_2)\geq e_1-\xi_1,\\
  & (Aw_2+e_1b_2)\leq e_1+\frac{1}{\tau}\xi_1,\nonumber
  \end{eqnarray} 
  where $c_1,~c_2 > 0$ are trade-off parameters and $e_1$ and $e_2$ are vectors of ones of appropriate dimensions. Notice that, similar to Pin-SVM, when $\tau=0$, the second constraint of both the problems become $\xi_1\geq 0,~\xi_2\geq 0$ and equation (\ref{eq:Pin-TWSVM1}) and equation (\ref{eq:Pin-TWSVM2}) respectively reduces to the traditional TWSVM.
 
Using Wolfe's dual method \cite{mangasarian1994nonlinear} and the K.K.T. \cite{mangasarian1994nonlinear},  Mehrkanoon et al. \cite{mehrkanoon2014non} obtained the dual of equation (\ref{eq:Pin-TWSVM1}) as follows:
 
  \begin{eqnarray}
  \label{eq:Dual-Pin-TWSVM1}
  \underset{(\alpha-\beta)}{Max}& e_2^T(\alpha-\beta)-\frac{1}{2}(\alpha-\beta)^TG^T(H^TH)^{-1}G(\alpha-\beta)\nonumber\\
  s.t. & c_1-\alpha-\frac{\beta}{\tau}=0,\\
  & \alpha\geq 0, ~\beta\geq 0,\nonumber
  \end{eqnarray}
  where $\alpha=(\alpha_1,\alpha_2,...,\alpha_{l_1})^T$ and $\beta=(\beta_1,\beta_2,...,\beta_{l_2})^T$ are the vectors of Lagrange multipliers. Here, $H=[A ~e_1]$ and $G=[B~e_2]$. Similarly, the dual problem of equation (\ref{eq:Pin-TWSVM2}) can be derived as follows:
    \begin{eqnarray}
    \label{eq:Dual-Pin-TWSVM2}
    \underset{(\gamma-\rho)}{Max}& e_1^T(\gamma-\rho)-\frac{1}{2}(\gamma-\rho)^TH^T(G^TG)^{-1}H(\gamma-\rho)\nonumber\\
    s.t. & c_2-\gamma-\frac{\rho}{\tau}=0,\\
    & \gamma\geq 0, ~\rho\geq 0.\nonumber
    \end{eqnarray}
    
    Here, the augmented vectors $u=[w_1,b_1]$ and $v=[w_2,b_2]$ are given by:
    \begin{eqnarray}
   u=(H^TH)^{-1}G^T(\alpha-\beta)\\
    v=(G^TG)^{-1}H^T(\gamma-\rho).
      \end{eqnarray} 
  Once vectors $u$ and $v$ are known from equation (7) and (8), the separating planes    
  \begin{eqnarray}
  x^Tw_1+b_1=0,~ ~~x^Tw_2+b_2=0,
  \end{eqnarray}
  are obtained. A new data sample $x_{new}\in \mathbb{R}^n$ is assigned to the class $1$ or class $-1$, based on which of the two hyperplanes it lies closest to, i.e.
  \begin{eqnarray}
  f(x_{new})=argmin\{d_{1,2}(x_{new})\},
  \end{eqnarray}
  where
  \begin{eqnarray}
  d_{1,2}=\left|{\frac{x_{new}^Tw_{1,2}+b_{1,2}}{||w_{1,2}||^2_2}}\right|\end{eqnarray}
  where $|\cdot|$ is the absolute value of distance of point $x_{new}$ from the plane.

\section{Novel Twin Support Vector Machine with Pinball Loss function based on Privilege Information}
  In this section, we introduce a novel approach of classification termed as Twin Support Vector Machine with Pinball Loss based on Privilege Information (Pin-TWSVMPI). In \cite{qi2014new}, Qi et al. have solved two additional QPPs model to incorporate privileged information in TWSVM which further leads to the higher computational cost of the model. In order to avoid, solving two additional QPPs model, in this paper, we utilize the novel method proposed in which extract privileged information from the dataset and then incorporate the same to the proposed Pin-TWSVMPI model. This leads to the lesser computational cost of proposed Pin-TWSVMPI as compared to FTWSVMPI. 
  
\begin{figure}[htp]
                     \centering
                     \includegraphics[width = 2.8in,height=2.5in]{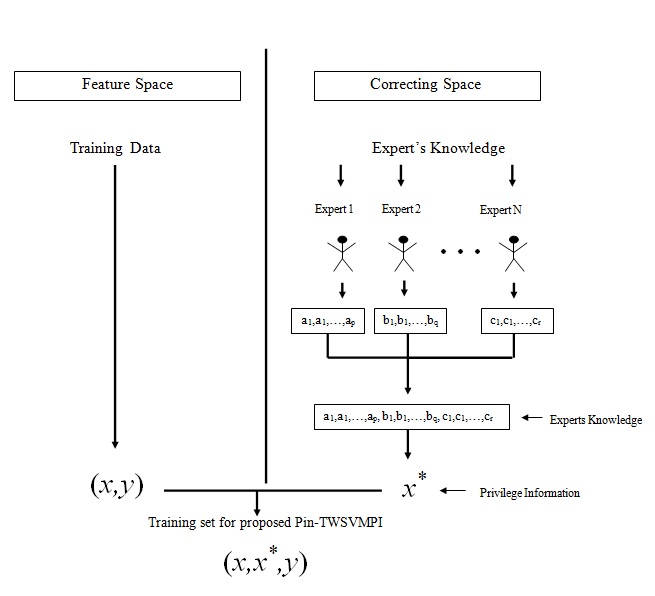} 
                     \caption{Process to generate privileged information from different experts knowledge } 
                     \label{Fig:Process of PI}
               \end{figure}

  The process for generating privileged information from the different experts is shown in fig. \ref{Fig:Process of PI} which further used in training the proposed Pin-TWSVMPI classifier. This figure also shows that an expert always supplies privileged information $x^*$ in the correcting space of $X^*$ with the admissible set of the correcting functions \cite{vapnik2009new}.
    
  \subsection{Linear Pin-TWSVMPI}
  Once we have generated privileged information from different experts, we map input feature vector $x$ of our training triplets $(x,x^*,y)$ into space $X$ and privileged information vector $x^*$ into $X^*$ where we define our decision rule as $(w^TX+b)$ and correcting function as $\xi =(w^{*T}x^*+b^*)$ in proposed Pin-TWSVMPI model. To find these functions we minimize the following QPPs:
  \begin{eqnarray}
  \label{eq:P-Pin-TWSVMPI1}
  \underset{w_1,\;b_1,\;w_1^*,\;b_1^*}{Min} & \frac{1}{2}||w_1||_2^2+\frac{\gamma}{2}||w_1^*||^2_2+\frac{1}{2}||\eta_1||^2_2+\frac{\gamma}{2}||\eta_1^*||_2^2+c_1e_2^{*T}(B^*w_1^*+e_2^*b_1^*)\nonumber\\
  s.t. & Aw_1+e_1b_1=\eta_1\\
  & A^*w_1^*+e_1^*b_1^*=\eta_1^*\nonumber\\
  & -(Bw_1+e_2b_1)\geq e_2-(B^*w_1^*+e_2^*b_1^*),\nonumber\\
  & -(Bw_1+e_2b_1)\leq e_2+\frac{1}{\tau}(B^*w_1^*+e_2^*b_1^*),\nonumber
  \end{eqnarray}  
 and
 \begin{eqnarray}
 \label{eq:P-Pin-TWSVMPI2}
   \underset{w_2,\;b_2,\;w_2^*,\;b_2^*}{Min} & \frac{1}{2}||w_2||_2^2+\frac{\gamma}{2}||w_2^*||^2_2+\frac{1}{2}||\eta_2||^2_2+\frac{\gamma}{2}||\eta_2^*||_2^2+c_2e_1^{*T}(A^*w_2^*+e_1^*b_2^*)\nonumber\\
   s.t. & Bw_2+e_2b_2=\eta_2\\
   & B^*w_2^*+e_2^*b_2^*=\eta_2^*\nonumber\\
   & (Aw_2+e_1b_2)\geq e_1-(A^*w_2^*+e_1^*b_2^*),\nonumber\\
   & (Aw_2+e_1b_2)\leq e_1+\frac{1}{\tau}(A^*w_2^*+e_1^*b_2^*),\nonumber
   \end{eqnarray}   
where $\gamma>0$ is a parameter. 

The proposed Pin-TWSVMPI implements the Structured Risk Minimization (SRM) principle \cite{gunn1998support} represented by first two terms of the objective function (\ref{eq:P-Pin-TWSVMPI1}) in feature space and correcting space respectively, which ensure for good generalization ability. The third and fourth terms of the objective function (\ref{eq:P-Pin-TWSVMPI1}) minimizes the projection of the class A samples in feature space and correcting space respectively, from the desired hyperplane, i.e., it tries to drive the samples of class A close to the desired hyperplane which is representation of class A. Since the proposed Pin-TWSVMPI is formulated as a soft-margin classifier, it permits violations of constraints represented by the fifth term of the objective function of (\ref{eq:P-Pin-TWSVMPI1}). The proposed Pin-TWSVMPI takes into consideration the principle of empirical risk minimization (ERM). The last constraints of (\ref{eq:P-Pin-TWSVMPI1}) implements the pinball loss function. Similar illustration can be drawn for (\ref{eq:P-Pin-TWSVMPI2}).

The dual problem of above discussed primal problem can be derived by using wolfe's dual method.  The Lagrangian function corresponding to the problem equation (\ref{eq:P-Pin-TWSVMPI1}) is given by
\begin{eqnarray}
\label{eq:Lag-Pin-TWSVMPI}
L(w_1,b_1,w_1^*,b_1^*,\eta_1,\eta_1^*,\alpha,\beta)=& \frac{1}{2}||w_1||_2^2+\frac{\gamma}{2}||w_1^*||^2_2+\frac{1}{2}||\eta_1||^2_2+\frac{\gamma}{2}||\eta_1^*||_2^2\nonumber\\
& +c_1e_2^{*T}(B^*w_1^*+e_2^*b_1^*)+\alpha_1^T[Aw_1+e_1b_1-\eta_1]\nonumber\\
& +\alpha_2^T[A^*w_1^*+e_1^*b_1^*-\eta_1^*]-\alpha_3^T[-(Bw_1+e_2b_1)\nonumber\\
& -e_2+(B^*w_1^*+e_2b_1^*)]+\alpha_4^T[-(Bw_1+e_2b_1)\nonumber\\
& -e_2+\frac{1}{\tau}(B^*w_1^*+e_2b_1^*)],
\end{eqnarray}
 where $\alpha_1,\alpha_2,\alpha_3,\alpha_4$ are vectors of Lagrange multipliers with length $l$. The Karush-Kuhn-Tucker (KKT) necessary and sufficient optimality conditions \cite{mangasarian1994nonlinear} for equation (\ref{eq:Lag-Pin-TWSVMPI}) are given by
 
 \begin{eqnarray}
 w_1+A^T\alpha_1+B^T\alpha_3-B^T\alpha_4 &=& 0,\\
 e_1^T\alpha_1+e_2^T\alpha_3-e_2\alpha_4 & =& 0,
 \end{eqnarray}
  \begin{eqnarray}
 \gamma w_1^*+c_1B^{*T}e_2^*+A^{*T}\alpha_2-B^{*T}\alpha_3-\frac{1}{\tau}B^{*T}\alpha_4 &=& 0,\\
  c_1 e_2^{*T}e_2^*+e_1^{*T}\alpha_2-e_2^{*T}\alpha_3-\frac{1}{\tau}e_2^{*T}\alpha_4 &=& 0,\\
  \eta_1-\alpha_1 &=& 0,\\
  \eta_1^*-\alpha_2&=& 0,\\
  Aw_1+e_1b_1&=& \eta_1,\\
  A^*w_1^*+e_1^*b_1^*&=& \eta_1^*,\\
  -(Bw_1+e_2b_1)+(B^*w_1^*+e_2^*b_1^*)  &\geq& e_2,\\
  e_2+(Bw_1+e_2b_1)-\frac{1}{\tau}(B^*w_1^*+e_2^*b_1^*)&\geq& 0,\\
  \alpha_1[Aw_1+e_1b_1-\eta_1] &=& 0,\\
  \alpha_2[A^*w_1^*+e_1^*b_1^*- \eta_1^*]&=& 0,\\
  \alpha_3[-(Bw_1+e_2b_1)-e_2+(B^*w_1^*+e_2^*b_1^*)] &=& 0,\\
  -\alpha_4[-(Bw_1+e_2b_1)-e_2+\frac{1}{\tau}(B^*w_1^*+e_2^*b_1^*)] &=& 0,\\
  \alpha_1 \geq~~ 0,~~\alpha_2\geq~~ 0,~~\alpha_3\geq~~ 0,~~\alpha_4 &\geq& 0.    
 \end{eqnarray}
 
 Now, from (14) and (16), we get, 
   \begin{eqnarray}
   \label{eq:u-pin-TWSVMPI}
   w_1=B^T(\alpha_4-\alpha_3)-A^T\alpha_1,
   \end{eqnarray}
and   
    \begin{eqnarray}
    \label{eq:u*-pin-TWSVMPI}
       w_1^*=\frac{1}{\gamma}\bigg[B^{*T}(\alpha_3+\frac{1}{\tau}\alpha_4-c_1e_2^*)-A^{*T}\alpha_2\bigg], 
       \end{eqnarray}
       respectively. Using equation (\ref{eq:Lag-Pin-TWSVMPI}) and the above K.K.T conditions, we obtain the Wolfe dual of equation (\ref{eq:P-Pin-TWSVMPI1}) as follows:
\begin{eqnarray}
\label{eq:D-Pin-TWSVMPI1}
\underset{\alpha_1,\alpha_2,\alpha_3,\alpha_4}{Min}& \frac{1}{2}(\alpha_4-\alpha_3)^TBB^T(\alpha_4-\alpha_3)+\frac{1}{2}\alpha_1AA^T\alpha_1+(\alpha_4-\alpha_3)BA^T\alpha_1\nonumber\\
& +\frac{1}{2\gamma}(\alpha_3+\frac{1}{\tau}\alpha_4-c_1e_2^*)^TB^*B^{*T}(\alpha_3+\frac{1}{\tau}\alpha_4-c_1e_2^*) +\frac{1}{2\gamma}\alpha_2^TA^*A^{*T}\alpha_2\nonumber\\
&-\frac{1}{\gamma}(\alpha_3+\frac{1}{\tau}\alpha_4-c_1e_2^*)^TB^*A^{*T}\alpha_2+e_2^T(\alpha_3-\alpha_2)\nonumber\\  
         s.t. &\\
                  	       & e_1^T\alpha_1-e_2^T(\alpha_4-\alpha_3) = 0,\nonumber\\
                  	       & e_1^{*T}\alpha_2-(\alpha_3+\frac{1}{\tau}\alpha_4-c_1e_2^*)^Te_2^*= 0,\nonumber\\
                  	       & \alpha_1,\alpha_2,\alpha_3,\alpha_4\geq 0.\nonumber         
         \end{eqnarray}
Next, from (18) and (20), we get,
\begin{eqnarray}
b_1=\frac{1}{l}(\alpha_1-Aw_1).
\end{eqnarray}         
Similarly from (19) and (21), we get
\begin{eqnarray}
b_1^*=\frac{1}{l}(\alpha_2-A^*w_1^*).
\end{eqnarray}

Furthermore let,
\begin{eqnarray}
Q=\begin{bmatrix}
AA^T & 0 & -AB^T & AB^T\\[4pt]
0 & \frac{1}{\gamma}A^*A^{*T} & -\frac{1}{\gamma}A^*B^{*T} & -\frac{1}{\gamma\tau}A^*B^{*T}\\[4pt]
-BA^T & -\frac{1}{\gamma}B^*A^{*T} & BB^T+\frac{1}{\gamma}B^*B^{*T} & -BB^T+\frac{1}{\gamma\tau}B^*B^{*T}\\[4pt]
BA^T & -\frac{1}{\gamma\tau}B^*A^{*T} & -BB^T+\frac{1}{\gamma\tau}B^*B^{*T} & BB^T+\frac{1}{\gamma\tau}B^*B^*T
\end{bmatrix},\nonumber
\end{eqnarray}         
\begin{eqnarray}
f=\begin{bmatrix}
0 & (c_1e_2^{*T}B^*A^{*T}-e_2) & (e_2-c_1\frac{2}{\gamma}B^*B^{*T}) & c_1\frac{2}{\gamma\tau}e_2^{*T}B^*B^{*T}
\end{bmatrix},\nonumber
\end{eqnarray}
\begin{eqnarray}
C=\begin{bmatrix}
e_1 & 0 & e_2 &-e_2\\
0 & e_1^* & -e_2^* & -\frac{1}{\tau}e_2^*
\end{bmatrix},~~~D=\begin{bmatrix}
0 & c_1
\end{bmatrix}^T,\nonumber
\end{eqnarray}       
and 
\begin{eqnarray}
X=\begin{bmatrix}
\alpha_1 & \alpha_2 & \alpha_3 & \alpha_4
\end{bmatrix}^T.\nonumber
\end{eqnarray}  

Thus, the problem (\ref{eq:D-Pin-TWSVMPI1}) can be reduced to
\begin{eqnarray}
\label{Eq:Proposed_formulation_SMO_form1}
\underset{X}{min} & \frac{1}{2}X^TQX+f^TX\nonumber\\
s.t. & CX=D\\
     & X\geq 0.   \nonumber 
\end{eqnarray}

The problem (\ref{Eq:Proposed_formulation_SMO_form1}) is a convex QPP and has formulation similar to the dual of SVM \cite{vapnik1998statistical}. Therefore, the well known Sequential Minimal Optimization (SMO) \cite{platt1998sequential} could used to solve the aforementioned problem.

Similarly, the dual problem of equation (\ref{eq:P-Pin-TWSVMPI2}) can also be derived as follows:
          \begin{eqnarray}
          \label{eq:D-Pin-TWSVMPI2}
\underset{\beta_1,\beta_2,\beta_3,\beta_4}{Min}& \frac{1}{2}(\beta_4-\beta_3)^TAA^T(\beta_4-\beta_3)+\frac{1}{2}\beta_1BB^T\beta_1+(\beta_4-\beta_3)AB^T\beta_1\nonumber\\
& +\frac{1}{2\gamma}(\beta_3+\frac{1}{\tau}\beta_4-c_2e_1^*)^TA^*A^{*T}(\beta_3+\frac{1}{\tau}\beta_4-c_2e_1^*) +\frac{1}{2\gamma}\beta_2^TB^*B^{*T}\beta_2\nonumber\\
&-\frac{1}{\gamma}(\beta_3+\frac{1}{\tau}\beta_4-c_2e_1^*)^TA^*B^{*T}\beta_2+e_1^T(\beta_3-\beta_2)\nonumber\\  
         s.t. &\\
                  	       & e_2^T\beta_1-e_1^T(\beta_4-\beta_3) = 0,\nonumber\\
                  	       & e_2^{*T}\beta_2-(\beta_3+\frac{1}{\tau}\beta_4-c_2e_1^*)^Te_1^*= 0,\nonumber\\
                  	       & \beta_1,\beta_2,\beta_3,\beta_4\geq 0.\nonumber
                   \end{eqnarray}
 Here, $(w_2,~b_2)$ and $(w_2^*,~b_2^*)$ can be calculated as 
    \begin{eqnarray}
    \label{eq:v-pin-TWSVMPI}
 w_2&=& A^T(\beta_4-\beta_3)-B^T\beta_1,\nonumber\\
 b_2&=& \frac{1}{l}(\beta_1-Bw_2). \nonumber
      \end{eqnarray} 
      and
      \begin{eqnarray}
      \label{eq:v*-pin-TWSVMPI}
      w_2^*&=&\frac{1}{\gamma}\bigg[A^{*T}(\beta_3+\frac{1}{\tau}\beta_4-c_2e_1^*)-B^{*T}\beta_2\bigg],\nonumber\\
      b_2^*&=&\frac{1}{l}(\beta_2-B^*w_2^*).
      \end{eqnarray}
         
Similar to (\ref{eq:D-Pin-TWSVMPI1}), SMO can also be used to solve (\ref{eq:D-Pin-TWSVMPI2}).

 Once vectors $w_1,~w_2,~w_1^*,~w_2^*$ and scalars $b_1,b_2,b_1^*,b_2^*$ are obtained from above, the separating hyperplanes 
 \begin{eqnarray}
   x^Tw_1+b_1=0,~` \mbox{and} ~~x^Tw_2+b_2=0,
   \end{eqnarray}      
   and corresponding correcting functions
             \begin{eqnarray}
                x^{*T}w_1^*+b_1^*=0,~` \mbox{and} ~~x^{*T}w_2^*+b_2^*=0,
                \end{eqnarray}   
                are obtained.
Since privileged information is available only at the time of training, therefore a new data sample $x_{new}\in \mathbb{R}^n$ is assigned to the class $1$ or to class $-1$, based on which of the two hyperplanes it lies closest to, i.e.
  \begin{eqnarray}
  f(x_{new})=argmin\{d_{\pm}(x_{new})\},
  \end{eqnarray}
  where
  \begin{eqnarray}
  d_{\pm}(x_{new})=\left|{\frac{x_{new}^Tw_\pm+b_\pm}{||w_\pm||^2_2}}\right|,\end{eqnarray}
  where $|\cdot|$ is the absolute value of distance of point $x_{new}$ from the plane. 

  \subsection{Nonlinear Pin-TWSVMPI }
  The nonlinear Pin-TWSVMPI can be expressed by introducing kernel generated hyperplanes by
   \begin{eqnarray}
     K(A,X^T)\mu_1+e_1\nu_1=0~~\mbox{and}~~K(B,X^T)\mu_2+e_2\nu_2=0,
   \end{eqnarray}
  where $X=A\cup B$ and $K(x_i,x_j)=(\phi(x_i)\cdot\phi(x_j))$ is an appropriate chosen kernel. Now the kernel correcting function is defined by $\xi_1 =(K(A^*,X^*)^*\mu_1^{*}+\nu_1^*)$ and $\xi_2 =(K(B^*,X^*)^*\mu_2^{*}+\nu_2^*)$. Next, the optimization problems for nonlinear Pin-TWSVMPI can be expressed as
  \begin{eqnarray}
    \label{eq:nonP-Pin-TWSVMPI1}
    \underset{\mu_1,\;\nu_1,\;\mu_1^*,\;\nu_1^*}{Min} & \frac{1}{2}||\mu_1||_2^2+\frac{\gamma}{2}||\mu_1^*||^2_2+\frac{1}{2}||\lambda_1||^2_2+\frac{\gamma}{2}||\lambda_1^*||_2^2+c_1e_2^{*T}(K(B^*,X^{*T})\mu_1^*+e_2^*\nu_1^*)\nonumber\\
      s.t. & \nonumber\\
      & K(A,X^T)\mu_1+e_1\nu_1=\lambda_1\nonumber\\
      & K(A^*,X^{*T})\mu_1^*+e_1^*\nu_1^*=\lambda_1^*\nonumber\\
       & -(K(B,X^T)\mu_1+e_2\nu_1)\geq e_2-(K(B^*,X^{*T})\mu_1^*+e_2^*\nu_1^*),\\
               & -(K(B,X^T)\mu_1+e_2\nu_1)\leq e_2+\frac{1}{\tau}(K(B^*,X^{*T})\mu_1^*+e_2^*\nu_1^*),\nonumber
    \end{eqnarray}  
                 and
                 \begin{eqnarray}
                 \label{eq:nonP-Pin-TWSVMPI2}
                 \underset{\mu_2,\;\nu_2,\;\mu_2^*,\;\nu_2^*}{Min} & \frac{1}{2}||\mu_2||_2^2+\frac{\gamma}{2}||\mu_2^*||^2_2+\frac{1}{2}||\lambda_2||^2_2+\frac{\gamma}{2}||\lambda_2^*||_2^2+c_2e_1^{*T}(K(A^*,X^{*T})\mu_2^*+e_1^*\nu_2^*)\nonumber\\
                    s.t. & \nonumber\\
                    & K(B,X^T)\mu_2+e_2\nu_2=\lambda_2\nonumber\\
                    & K(B^*,X^{*T})\mu_2^*+e_2^*\nu_2^*=\lambda_2^*\nonumber\\
                    & (K(A,X^T)\mu_2+e_1\nu_2)\geq e_1-(K(A^*,X^{*T})\mu_2^*+e_1^*\nu_2^*),\\
                    & (K(A,X^T)\mu_2+e_1\nu_2)\leq e_1+\frac{1}{\tau}(K(A^*,X^{*T})\mu_2^*+e_1^*\nu_2^*).\nonumber
                    \end{eqnarray}

Similar to the linear case, the solution of equation (\ref{eq:nonP-Pin-TWSVMPI1}) and equation (\ref{eq:nonP-Pin-TWSVMPI2}) can be derived from wolfe's dual method as follows:
\begin{eqnarray}
         \label{eq:nonD-Pin-TWSVMPI1}
         \underset{\alpha_1,\alpha_2,\alpha_3,\alpha_4}{Min}& \frac{1}{2}(\alpha_4-\alpha_3)^TNN^T(\alpha_4-\alpha_3)+\frac{1}{2}\alpha_1MM^T\alpha_1+(\alpha_4-\alpha_3)NM^T\alpha_1\nonumber\\
         & +\frac{1}{2\gamma}(\alpha_3+\frac{1}{\tau}\alpha_4-c_1e_2^*)^TN^*N^{*T}(\alpha_3+\frac{1}{\tau}\alpha_4-c_1e_2^*) +\frac{1}{2\gamma}\alpha_2^TM^*M^{*T}\alpha_2\nonumber\\
         &-\frac{1}{\gamma}(\alpha_3+\frac{1}{\tau}\alpha_4-c_1e_2^*)^TN^*M^{*T}\alpha_2+e_2^T(\alpha_3-\alpha_2)\nonumber\\  
                  s.t. &\\
                           	       & e_1^T\alpha_1-e_2^T(\alpha_4-\alpha_3) = 0,\nonumber\\
                           	       & e_1^{*T}\alpha_2-(\alpha_3+\frac{1}{\tau}\alpha_4-c_1e_2^*)^Te_2^*= 0,\nonumber\\
                           	       & \alpha_1,\alpha_2,\alpha_3,\alpha_4\geq 0.\nonumber 
          \end{eqnarray}
                   and
                   
  \begin{eqnarray}
   \label{eq:nonD-Pin-TWSVMPI2}
   \underset{\beta_1,\beta_2,\beta_3,\beta_4}{Min}& \frac{1}{2}(\beta_4-\beta_3)^TMM^T(\beta_4-\beta_3)+\frac{1}{2}\beta_1NN^T\beta_1+(\beta_4-\beta_3)MN^T\beta_1\nonumber\\
   & +\frac{1}{2\gamma}(\beta_3+\frac{1}{\tau}\beta_4-c_2e_1^*)^TM^*M^{*T}(\beta_3+\frac{1}{\tau}\beta_4-c_2e_1^*) +\frac{1}{2\gamma}\beta_2^TN^*N^{*T}\beta_2\nonumber\\
   &-\frac{1}{\gamma}(\beta_3+\frac{1}{\tau}\beta_4-c_2e_1^*)^TM^*N^{*T}\beta_2+e_1^T(\beta_3-\beta_2)\nonumber\\  
            s.t. &\\
                     	       & e_2^T\beta_1-e_1^T(\beta_4-\beta_3) = 0,\nonumber\\
                     	       & e_2^{*T}\beta_2-(\beta_3+\frac{1}{\tau}\beta_4-c_2e_1^*)^Te_1^*= 0,\nonumber\\
                     	       & \beta_1,\beta_2,\beta_3,\beta_4\geq 0.\nonumber
                    \end{eqnarray}
where $M=[K(A,X^T)~e_1]$, $N=[K(B,X^T)~e_2]$, $M^*=[K(A^*,X^{*T})~e_1^*]$, and $N^*=[K(B^*,X^{*T})~e_2^*]$.  

Similar to linear case, SMO can be used to solve equation (\ref{eq:nonD-Pin-TWSVMPI1}) and equation (\ref{eq:nonD-Pin-TWSVMPI2}) respectively. Once the vector of dual variables $\alpha_1,\alpha_2,\alpha_3,\alpha_4$ and $\beta_1,\beta_2,\beta_3,\beta_4$ are determined, the vector of primal variables are obtained as follows:
\begin{eqnarray}
    \label{eq:nonu-pin-TWSVMPI}
\mu_1&=& N^T(\alpha_4-\alpha_3)-M^T\alpha_1,\nonumber\\
\nu_1&=& \frac{1}{l}(\alpha_1-M\mu_1),\nonumber\\
\mu_1^*&=& \frac{1}{\gamma}\bigg[\alpha_3+\frac{1}{\tau}\alpha_4-c_1N^{*T}1e_2^*-M^{*T}\alpha_2\bigg],\nonumber\\
\nu_1^*&=&\frac{1}{l}(\alpha_2-A^*w_1^*),\nonumber 
\end{eqnarray}
and
\begin{eqnarray}
\label{eq:nonu*-pin-TWSVMPI}
 \mu_2&=& M^T(\beta_4-\beta_3)-N^T\beta_1,\nonumber\\
 \nu_2&=& \frac{1}{l}(\beta_1-N\mu_2), \nonumber\\
 \mu_2^*&=&\frac{1}{\gamma}\bigg[\beta_3+\frac{1}{\tau}\beta_4-c_2M^{*T}e_1^*-N^{*T}\beta_2\bigg],\nonumber\\
\nu_2^*&=&\frac{1}{l}(\beta_2-N^*\mu_2^*).\nonumber
\end{eqnarray}           
      
      A new data sample $x_{new}\in \mathbb{R}^n$ is assigned to the class $1$ or to class $-1$, based on which of the two hyperplanes it lies closest to, i.e.
        \begin{eqnarray}
        f(x_{new})=argmin\{d_{1,2}(x_{new}))\},
        \end{eqnarray}
        where
        \begin{eqnarray}
        d_{1,2}(x_{new})=\left|{\frac{K(x_{new},X^T)^T\mu_{1,2}+\nu_{1,2}}{||\mu_{1,2}||^2_2}}\right|,\end{eqnarray}
        where $|\cdot|$ is the absolute value of distance of point $x_{new}$ from the plane.


\section{Two Applications of Privileged Information}
In this section, we discuss privileged information for pedestrian detection and handwritten digit recognition.

\subsection{Pedestrian Detection}
Pedestrian detection is a key problem in computer vision field \cite{min2018recognition}. In the last few years, diverse efforts have been made to improve the performance of pedestrian detection in \cite{viola2001rapid,zhang2015exploring,zhang2015filtered,paisitkriangkrai2014strengthening}.
In this paper, we also propose a fast and novel framework for pedestrian detection problem and explain each step involved therein. We use Pin-TWSVMPI which helps to enhance the performance of proposed pedestrian detection framework. The training dataset in pedestrian detection includes positive and negative images. Positive image means atleast one pedestrian is present in the image whereas negative image means pedestrian is not present in the image. In this framework, we extract HOG \cite{dalal2005histograms}, HOF \cite{park2013exploring} and gray level co-occurrence matrix \cite{haralick1973textural} features from the images. Here, each feature vector gives the different independent view of the same data, e.g., HOG represents training data view, HOF and gray level co-occurrence represents privileged information view. From another perspective, we may also term this privileged information view as different \lq expert's knowledge \rq which is extracted in the correcting space. Note that, expert's knowledge is only available at the time of training. Thus, for the testing image, we extract only HOG feature as the testing feature. The proposed framework would follow the following steps to detect pedestrians: 

\fbox{\begin{minipage}{30em}
\begin{itemize}
\item[]
\textbf{Training phase:} Input the training dataset (although this framework could be extended in general for any type of dataset), k-fold parameter, kernel parameter, $\tau$, $\gamma$, $c_1$ and $c_2$ parameters of Pin-TWSVMPI.
\item[] \textbf{Step 1:} Use Viola Jones method \cite{viola2001rapid} to identify the detection window for every pedestrian present in the positive image.  
\item[] \textbf{Step 2:} Each detection window is expressed by overlapping blocks, and we extract HOG in feature space and HOF and gray level co-occurrence matrix features in correcting space for each block.
\item[] \textbf{Step 3:}  Train Pin-TWSVMPI with the combination of extracted features and privileged information via $k$-fold cross validation strategy and obtain the optimal classifier.
\item[] \textbf{Testing phase:} Identify the pedestrian in a testing image i.e., either the pedestrian is presented or not in the given image with the help of trained Pin-TWSVMPI.
\end{itemize}
\end{minipage}}
\vspace{0.4cm}

\begin{figure}[htp]
                     \centering
                     \includegraphics[width = 4.5in,height=6.5in]{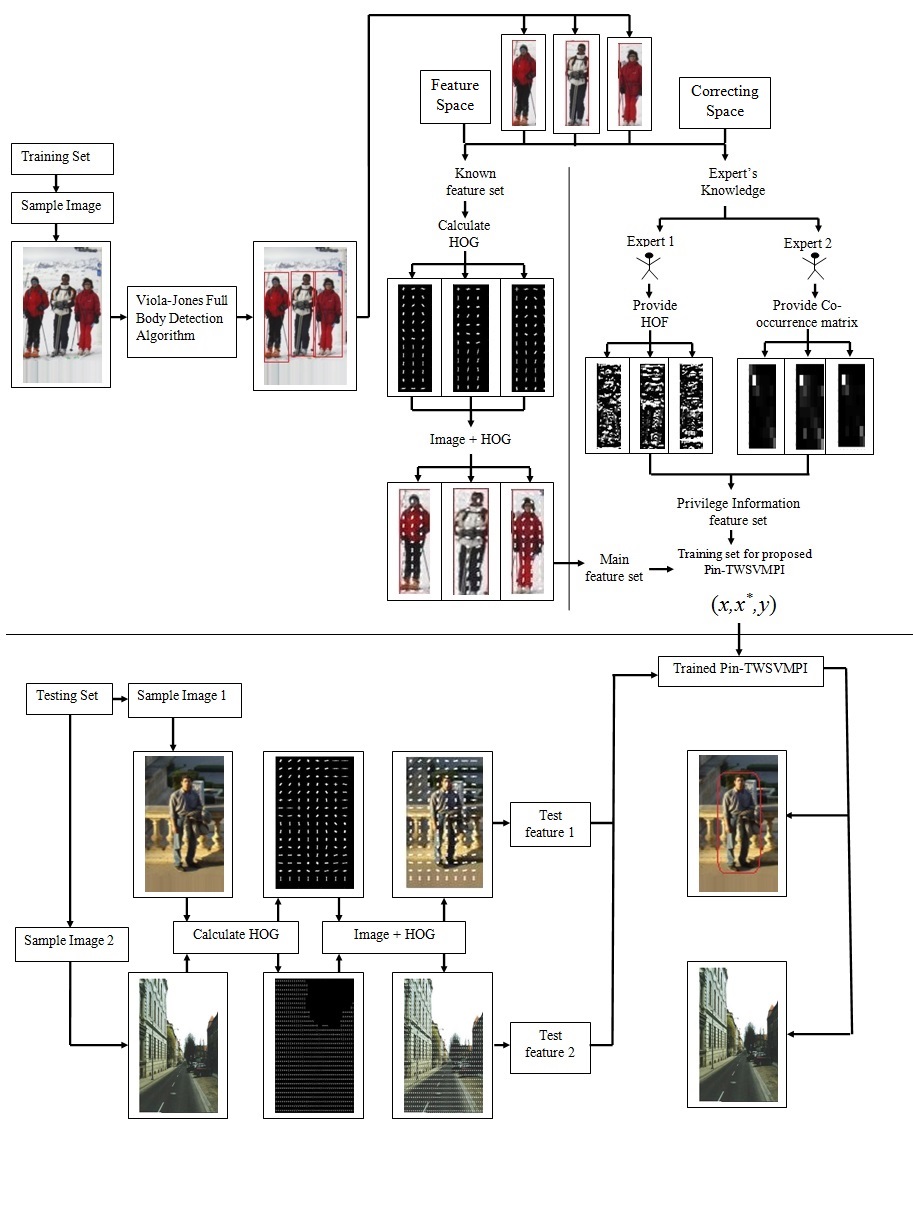} 
                     \caption{Complete procedure for pedestrian detection with proposed Pin-TWSVMPI} 
                     \label{Fig:Process of PI_ped_det}
\end{figure}

Figure \ref{Fig:Process of PI_ped_det} represents the graphical interpretation of the proposed framework to detect the pedestrian. In figure \ref{Fig:Process of PI_ped_det}, we have shown the result for both presence and absence pedestrian/object under consideration. This figure also shows that if the testing image does not have any pedestrian present in the image, then classifier will not identify the same. Note that, in our experiments, we have used only two expert knowledge, but this can be increased as per an expert's availability.  

A major highlight of the above-discussed framework is the freedom to incorporate knowledge from different experts without increasing the training time of the propose Pin-TWSVMPI classifier. 

\subsection{Handwritten Digit Recognition}
Handwritten digit recognition is a current research area in Optical Character Recognition (OCR) applications and pattern classification \cite{leou11comparison}. The performance of OCR largely based on the classification/learning scheme \cite{liu2003handwritten}. In recent years, SVM based classifiers have attracted the attention of researcher in the field of handwritten digit recognition \cite{sharma2012handwritten,sadri2003application}. 

To show as an application of proposed Pin-TWSVMPI on handwritten digit recognition, we used MNIST dataset which has become a standard benchmark dataset for learning handwritten digit recognition \cite{MNIST_ref}. In this paper, we classify the images of digits 5 and 8 in the MNIST database. Similar to \cite{vapnik2009new}, for every training image of MNIST dataset, we utilized its holistic (poetic) description which further used as the privileged information to the proposed model. The poetic description of every digit of MNIST dataset is directly taken from \cite{vapnik2009new}. These poetic descriptions for all training images were available in $21$-dimensional feature vectors which will act like the privileged information to the proposed model. The MNIST data and its poetic descriptions with corresponding feature vectors, can be downloaded from {\color{blue}http://www.nec-labs.com/research/machine/ml\_website/department/softwa\\re/learning-with-teacher}.

        
\section{Experimental Results}
Experiments were performed in MATLAB version $8.1$ on a machine with $3.40$ GHz CPU and $16$ GB RAM under Microsoft Windows 64-bit operating system. To check the performance of proposed Pin-TWSVMPI, where privileged information is available i.e., pedestrian detection and handwritten digit recognition problems, and when privileged information is not available i.e., UCI machine learning repository \cite{asuncion2007uci} datasets. Note that the bold values in the experiment tables represents the best value along rows.

  \subsection{Performance on UCI Database}
     \label{Sec:novel_method}
  In this section, we present the performance of the proposed formulation over UCI datasets where privileged information is not available to establish the classification accuracy of the proposed Pin-TWSVMPI. Privilege information is the additional information about the datasets which further help us to improve the performance of the proposed Pin-TWSVMPI. However, in a few problems, where an expert's knowledge is not available, the following procedure is used to extract privileged information:
  
  \textbf{Procedure to extract Privilege Information:}
           In order to extract privileged information, we have implemented the idea of Aman and Reshma \cite{pal2018learning} which retrieve prominent features from the feature set of data. To achieve this, we use Principal Components Analysis (PCA) \cite{abdi2010principal} which is a statistical method for determining patterns in the data, and indicating the data in such a way so as to distinguish their similarities and differences. PCA uses an orthogonal transformation to convert a set of observations of possibly correlated variables into a set of values of linearly uncorrelated variables called principal components \cite{smith2002tutorial}. This transformation is defined in such a way that the first major component has the greatest possible deviation (that is, corresponds to as much variability in the data as possible). Therefore, we use principal components derived from PCA as privileged information in the proposed formulation. Table \ref{Tab:UCI_accuracy_linear} and \ref{Tab:UCI_accuracy_nonlinear} show that acquiring privileged information via PCA and incorporating the same in the proposed formulation is effective for linear and nonlinear kernel respectively.
         
  The classification performance of proposed Pin-TWSVMPI has been evaluated through average training accuracy and $F1$-measure \cite{lipton2014optimal} over UCI datasets. The average training accuracy can be calculated as
  \begin{eqnarray}
  Testing~Accuracy=\frac{\text{No. of correctly classified testing samples}}{\text{Total no. of samples in testing set}}.
  \label{testing_accuracy}
  \end{eqnarray}
  and $F1$-measure can be calculated as 
  \begin{eqnarray}
  F1-score=2*\frac{\text{Precision}*\text{Recall}}{\text{Precision}+\text{Recall}}.
  \label{F1-measure}
  \end{eqnarray}
  Generalization error was determined by following the standard 5-fold cross-validation methodology \cite{duda1973pattern}. Average Testing Accuracy of all formulations for 5-fold cross-validation was represented by ($mean\pm std$) in the following tables. Optimal value of $c_1$, $c_2$, $\gamma$, $\tau$, and kernel parameter $\sigma$ were obtained by using a tuning set comprising of 10 percent of the data set which is further sent back to the training dataset. 
  
  Table \ref{Tab:SummaryUCI} summarizes different types of real-world datasets from the UCI machine learning repository \cite{asuncion2007uci}. These datasets include Iris, Compound, Wine, Ecoli, Dermatology, Zoo, Haberman, Soybean and Page Block. The missing values in the datasets were replaced by the mean value of their corresponding column.
  \begin{table}
  \centering
  \captionof{table}{Summary of UCI datasets.}
  \begin{tabular}{p{2.5cm}|p{3cm}|p{3cm}|p{3cm}}
   \hline
    Dataset & {No. of instances} & {No. of Features} & {No. of classes} \\ 
    \hline         
   Iris        & 150 &  4 &  3  \\
   Compound    & 399 &  2 &  6  \\  
   Wine        & 178 & 13 &  3  \\
   Ecoli       & 327 &  7 &  5  \\
   Dermatology & 358 & 34 &  6 \\ 
   Zoo         & 101 & 17 & 7 \\
   Haberman    & 303 &  3 & 2 \\ 
   Soybean     &  47 & 35 & 4 \\ 
   Page Block     &  5473 & 10 & 5 \\
   \hline 
  \end{tabular}
  \label{Tab:SummaryUCI} 
  \end{table}
  
\begin{figure}[htp]
                       \centering
                       \includegraphics[width = 4.5in,height=2.5in]{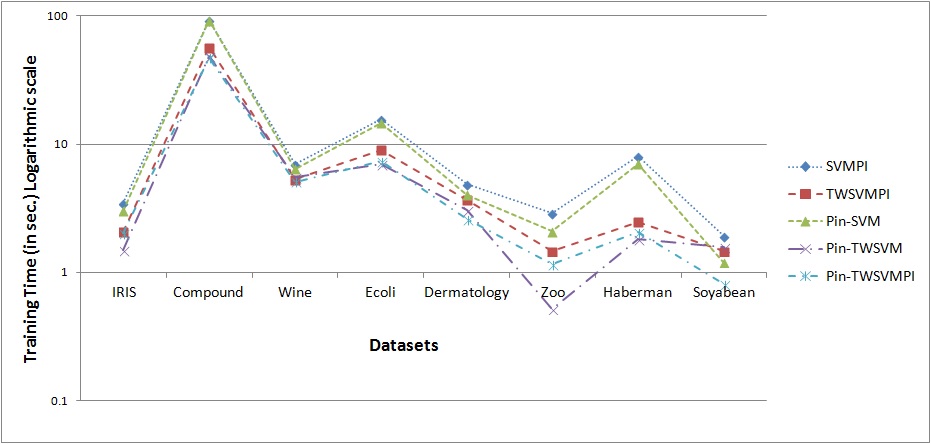} 
                       \caption{Comparison of training time with various methods} 
                       \label{Fig:Time_comp}
  \end{figure}

   Table \ref{Tab:UCI_accuracy_linear} and table \ref{Tab:UCI_accuracy_nonlinear} compares the classification accuracy of proposed classifier Pin-TWSVMPI with SVMPI, TWSVMPI, Pin-SVM, Pin-SVMPI and Pin-TWSVM with linear and nonlinear kernel respectively. Further, table \ref{Tab:UCI_accuracy_nonlinear} compares F1 score of  Pin-TWSVMPI and Pin-TWSVM. Figure \ref{Fig:Time_comp} shows the comparison of training time obtained by proposed Pin-TWSVMPI with SVMPI, TWSVMPI, Pin-SVM, Pin-SVMPI, and Pin-TWSVM. This figure shows that proposed Pin-TWSVMPI is faster among all. This is due to the SMO technique which is used to obtain the optimal classifier. 
   
   \begin{table}
   \centering
   \captionof{table}{Comparison of Average Testing Accuracy with standard deviation of different classifier on UCI datasets with linear kernel.}
   \begin{tabular}{ p{2.2cm}|p{2.2cm}|p{2.2cm}|p{2.2cm}|p{2.2cm}|p{2.2cm}}
    \hline
      \multirow{2}{*}{Dataset} & \multicolumn{5}{c}{ Classifiers} \\
      
      \cline{2-6}
            & SVMPI & TWSVMPI & Pin-SVM & Pin-TWSVM & Pin-TWSVMPI\\
    \hline
    Iris         & $95.65\pm 4.36$ & $95.66\pm 4.21$ & $95.89\pm 3.01$ & $96.24\pm 3.87$ & $\mathbf{98.01\pm 2.87}$ \\
    Compound  & $81.72\pm 4.07$ & $82.02\pm 5.05$ & $ 82.31\pm 4.87$ & $ 82.78\pm 5.41$ & $\mathbf{84.78\pm 3.66}$ \\
    Wine      & $95.45\pm 3.54$ & $96.00\pm 3.58$ & $96.12\pm 5.30 $ & $96.45\pm 4.65$ & $\mathbf{97.54\pm 3.86}$ \\
    Ecoli    & $82.28\pm 3.35$ &  $82.76\pm 5.87$ & $82.88\pm 4.77 $ & $83.04\pm 5.01$ & $\mathbf{85.68\pm 4.68}$ \\
    Dermatology  & $93.01\pm 3.76$ &  $94.52\pm 2.64$ & $94.21\pm 2.22$ & $94.04\pm 4.02$ & $\mathbf{95.98\pm 3.38}$ \\
    Zoo  & $91.65\pm 3.43$ & $92.00\pm 3.14$ & $92.41\pm 3.65$ & $93.00\pm 2.06$ & $\mathbf{94.31\pm 2.11}$ \\
    Glass  & $55.69\pm 4.11$ & $56.66\pm 5.07$ & $56.40\pm 4.78$ & $56.68\pm 4.54$ & $\mathbf{58.62\pm 2.94}$ \\
    Haberman  & $70.02\pm 4.12$ & $70.16\pm 4.89$ & $71.67\pm 3.36$ & $71.74\pm 5.04$ & $\mathbf{73.88\pm 3.03}$ \\
    Soybean  & $97.64\pm 1.08$ & $\mathbf{100.00\pm 0.00}$ & $98.86\pm 2.11$ & $\mathbf{100.00\pm 0.00}$ & $\mathbf{100.00\pm 0.00}$\\
    Page Block  & $88.14\pm 1.21$ & $90.47\pm 0.88 $ & $89.74\pm 1.34$ & $90.11\pm 1.23$ & $\mathbf{92.14\pm 0.21}$ \\
    \hline
   \end{tabular}
   \label{Tab:UCI_accuracy_linear} 
   \end{table}
   
  
  \begin{sidewaystable}
  \centering
  \captionof{table}{Comparison of Average Testing Accuracy with standard deviation of different classifier on UCI datasets with nonlinear kernel.}
  \begin{tabular}{p{2.2cm}|p{2.2cm}|p{2.2cm}|p{2.2cm}|p{2.2cm}|p{2.2cm}|p{2.2cm}}
   \hline
     \multirow{2}{*}{Dataset} & \multirow{2}{*}{Measures} & \multicolumn{5}{c}{ Classifiers} \\ \cline{3-7}
               & & SVMPI & TWSVMPI & Pin-SVM & Pin-TWSVM &  Pin-TWSVMPI\\
   \hline
   \multirow{2}{*}{Iris}         &  \text{Accuracy} & $96.61\pm 3.36$ & $96.66\pm 5.32$ & $96.78\pm 3.12$ & $97.04\pm 4.01$ & $\mathbf{98.85\pm 4.08}$  \\
   &  \text{F1 Score} & $-$ & $-$ & $-$ & $0.75$ & $\mathbf{0.80}$  \\
   \hline
   \multirow{2}{*}{Compound}  & Accuracy & $96.41\pm 3.98$ & $97.21\pm 5.74$ & $96.87\pm 5.78$ & $97.85\pm 5.14$ & $\mathbf{98.93\pm 3.08}$ \\
   &  \text{F1 Score} & $-$ & $-$ & $-$ & $0.75$ & $\mathbf{0.80}$  \\
   \hline
   \multirow{2}{*}{Wine}      & Accuracy & $98.12\pm 3.74$ & $98.85\pm 4.02$ & $98.45\pm 2.65$ & $98.81\pm 3.78$ & $\mathbf{98.96\pm 4.14}$ \\
   &  \text{F1 Score} & $-$ & $-$ & $-$ & $0.7105$ & $\mathbf{0.8840}$  \\
   \hline
   \multirow{2}{*}{Ecoli}    & Accuracy & $87.63\pm 4.87$ &  $88.00\pm 4.97$ & $87.94\pm 5.47$ & $88.50\pm 4.68$ & $\mathbf{90.11\pm 4.67}$ \\
   &  \text{F1 Score} & $-$ & $-$ & $-$ & $0.6042$ & $\mathbf{0.6291}$  \\
   \hline
   \multirow{2}{*}{Dermatology} & Accuracy & $94.89\pm 3.14$ &  $95.71\pm 2.78$ & $95.00\pm 2.74$ & $95.78\pm 3.47$ & $\mathbf{96.88\pm 2.09}$\\
   &  \text{F1 Score} & $-$ & $-$ & $-$ & $0.7584$ & $\mathbf{0.8045}$  \\
   \hline
   \multirow{2}{*}{Zoo}  & Accuracy & $94.00\pm 3.54$ & $95.00\pm 3.28$ & $94.15\pm 3.78$ & $95.45\pm 5.09$ & $\mathbf{96.79\pm 4.08}$ \\
   &  \text{F1 Score} & $-$ & $-$ & $-$ & $0.7202$ & $\mathbf{0.8211}$  \\
   \hline
   \multirow{2}{*}{Glass}  & Accuracy & $62.84\pm 3.12$ & $63.33\pm 4.47$ & $ 63.04\pm 2.68$ & $63.84\pm 4.65$ & $\mathbf{65.61\pm 3.88}$ \\
   &  \text{F1 Score} & $- Score$ & $-$ & $-$ & $0.5980$ & $\mathbf{0.6333}$  \\
   \hline
   \multirow{2}{*}{Haberman}  & Accuracy & $75.16\pm 4.08$ & $75.40\pm 5.21$ & $75.64\pm 4.98$ & $76.14\pm 6.06$ & $\mathbf{77.78 \pm 4.67}$\\
   &  \text{F1 Score}  & $-$ & $-$ & $-$ & $0.5043$ & $\mathbf{0.5539}$  \\
   \hline
   \multirow{2}{*}{Soybean}  & Accuracy & $\mathbf{100.00\pm 0.00}$ & $\mathbf{100.00\pm 0.00}$ & $\mathbf{100.00\pm 0.00}$ & $\mathbf{100.00 \pm 0.00}$ & $\mathbf{100.00\pm 0.00}$ \\
   &  \text{F1 Score} & $-$ & $-$ & $-$ & $-$ & $-$  \\
   \hline
   \multirow{2}{*}{Page Block}  & Accuracy & $91.76\pm 1.87$ & $92.74\pm 1.44$  & $90.75\pm 2.07$ & $92.74\pm 1.21$ & $\mathbf{94.45\pm 1.00}$ \\
   &  \text{F1 Score} & $-$ & $-$ & $-$ & $0.7986$ & $\mathbf{0.8655}$  \\
   \hline
  \end{tabular}
  \label{Tab:UCI_accuracy_nonlinear} 
  \end{sidewaystable}
  %
  
  \subsection{Performance on Pedestrian Detection}
\begin{figure}[htp]
   \centering
   \subfloat{(a)}~~~~\subfloat{\includegraphics[width = 0.7in,height=1in]{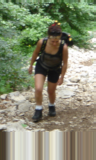}}~~~~
   \subfloat{\includegraphics[width = 0.7in,height=1in]{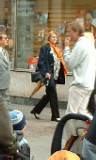}}~~~~
   \subfloat{\includegraphics[width = 0.7in,height=1in]{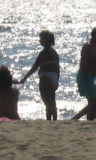}}~~~~
   \subfloat{\includegraphics[width = 0.7in,height=1in]{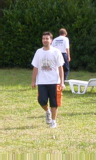}}~~~~
   \subfloat{\includegraphics[width = 0.7in,height=1in]{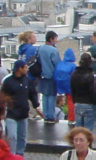}}~~~~
   \subfloat{\includegraphics[width = 0.7in,height=1in]{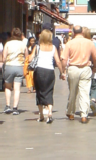}}  \\
   \subfloat{(b)}~~~~\subfloat{\includegraphics[width = 0.7in,height=1in]{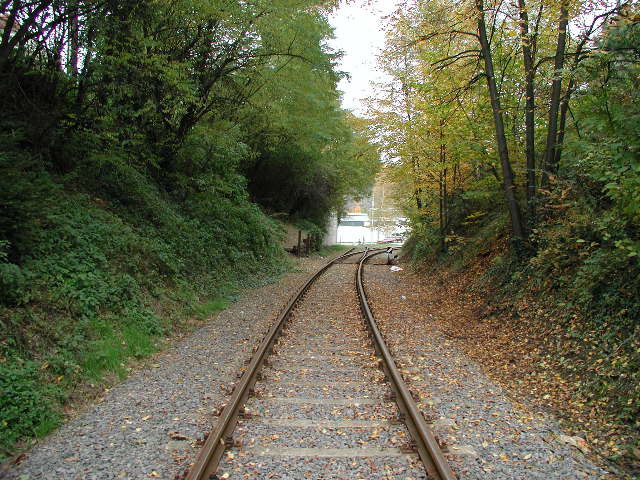}}~~~~
    \subfloat{\includegraphics[width = 0.7in,height=1in]{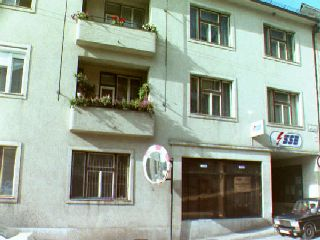}}~~~~
    \subfloat{\includegraphics[width = 0.7in,height=1in]{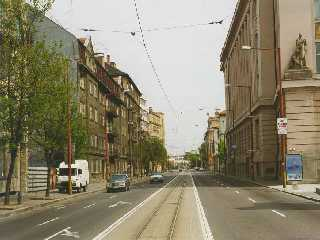}}~~~~
    \subfloat{\includegraphics[width = 0.7in,height=1in]{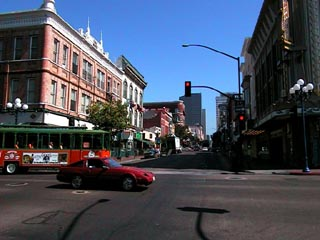}}~~~~
      \subfloat{\includegraphics[width = 0.7in,height=1in]{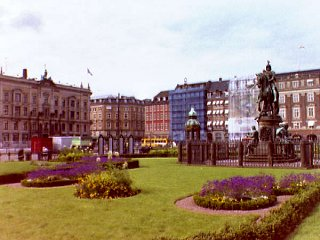}}~~~~
    \subfloat{\includegraphics[width = 0.7in,height=1in]{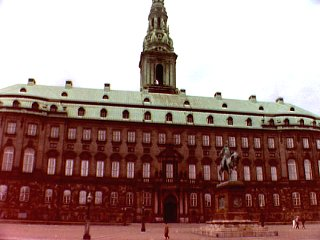}}  \\
    \subfloat{(c)}~~~~\subfloat{\includegraphics[width = 0.7in,height=1in]{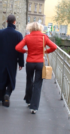}}~~~~
     \subfloat{\includegraphics[width = 0.7in,height=1in]{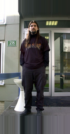}}~~~~
     \subfloat{\includegraphics[width = 0.7in,height=1in]{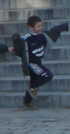}}~~~~
     \subfloat{\includegraphics[width = 0.7in,height=1in]{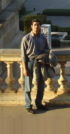}}~~~~ 
     \subfloat{\includegraphics[width = 0.7in,height=1in]{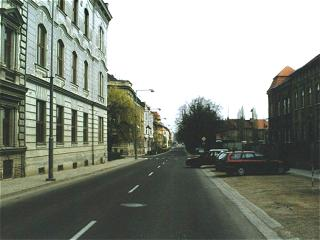}}~~~~
        \subfloat{\includegraphics[width = 0.7in,height=1in]{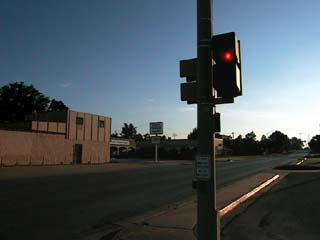}} 
   \caption{Sample images from benchmark INRIA dataset: (a) Pedestrian training samples, (b) Nonpedestrain training sapmles and (c) Test images.} 
   \label{Fig:sample_inria}
                 \end{figure}

  In this section, we used the standard INRIA dataset \cite{dalal2005histograms} to demonstrate the effectiveness of the pedestrian detector. INRIA dataset consists of 1805 pedestrian images with the size of $(64 \times 128)$. In the experiment, we choose HOG as the standard input data, and HOF and Co-occurrence matrix feature descriptors as the privileged information, respectively. Further, we choose 1271 pedestrians images as the positive training samples, and negative samples are chosen from the rest of the images, which do not contain pedestrians. Figure \ref{Fig:sample_inria} shows the sample images from INRIA dataset. 
    
\begin{figure}[htp]
                         \centering
                         \includegraphics[width = 4.5in,height=2.5in]{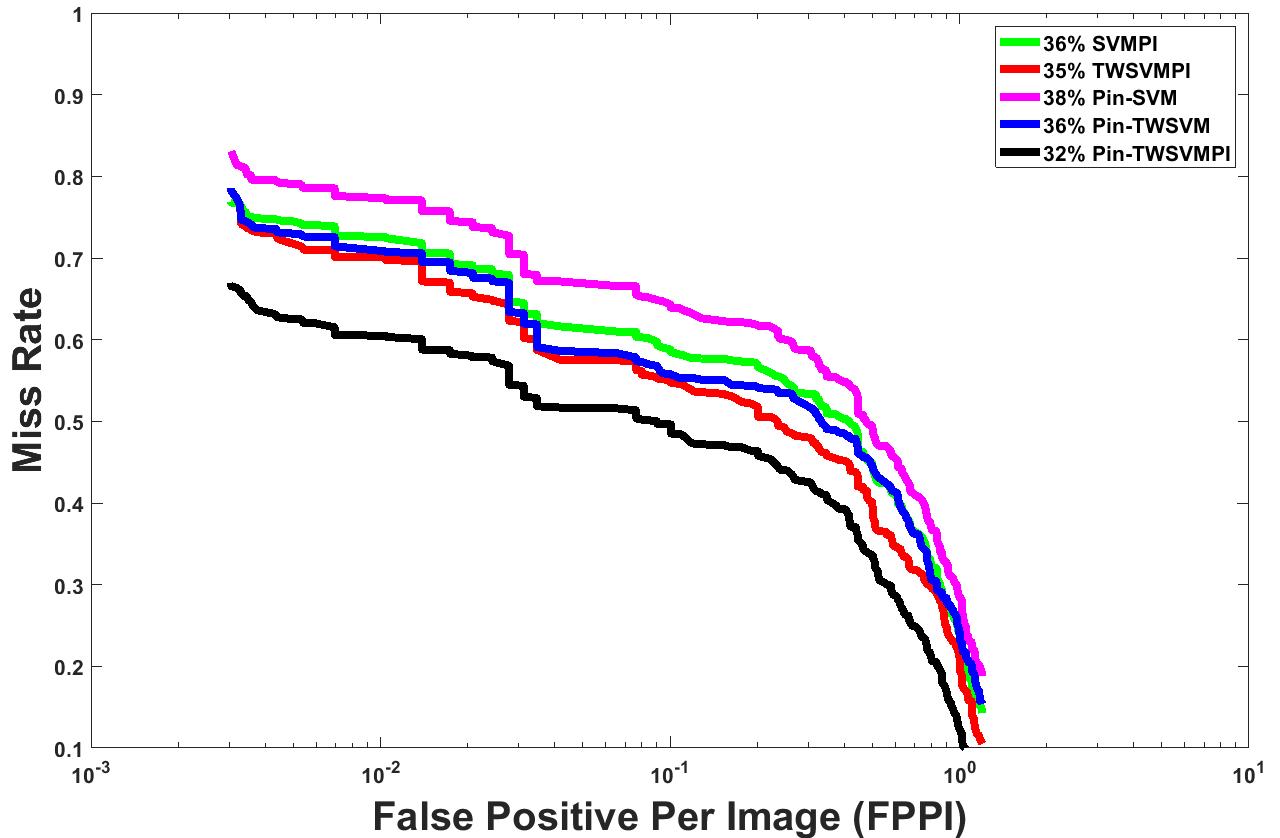} 
                         \caption{Comparison of pedestrian detection result of proposed Pin-TWSVMPI with SVMPI, TWSVMPI, Pin-SVM and Pin-TWSVM} 
                         \label{Fig:pedestrian_result}
    \end{figure}

\begin{figure}[htp]
    \centering
    \subfloat{(a)}~~~~\subfloat{\includegraphics[width = 0.8in,height=1in]{test1.jpg}}~~~~
    \subfloat{\includegraphics[width = 0.8in,height=1in]{test2.jpg}}~~~~
    \subfloat{\includegraphics[width = 0.8in,height=1in]{test3.jpg}}~~~~
    \subfloat{\includegraphics[width = 0.8in,height=1in]{test4.jpg}}~~~~
    \subfloat{\includegraphics[width = 0.8in,height=1in]{test5.jpg}}~~~~
    \subfloat{\includegraphics[width = 0.8in,height=1in]{test6.jpg}} \\ 
    \subfloat{(b)}~~~~\subfloat{\includegraphics[width = 0.8in,height=1in]{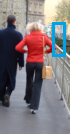}}~~~~
     \subfloat{\includegraphics[width = 0.8in,height=1in]{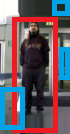}}~~~~
     \subfloat{\includegraphics[width = 0.8in,height=1in]{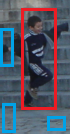}}~~~~
     \subfloat{\includegraphics[width = 0.8in,height=1in]{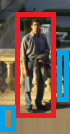}}~~~~
      \subfloat{\includegraphics[width = 0.8in,height=1in]{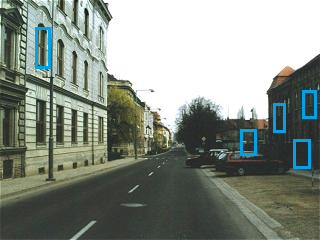}}~~~~
      \subfloat{\includegraphics[width = 0.8in,height=1in]{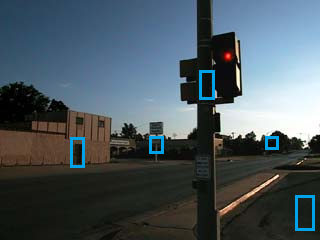}}\\
   \subfloat{(c)}~~~~\subfloat{\includegraphics[width = 0.8in,height=1in]{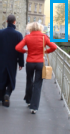}}~~~~
    \subfloat{\includegraphics[width = 0.8in,height=1in]{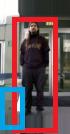}}~~~~
    \subfloat{\includegraphics[width = 0.8in,height=1in]{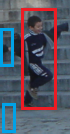}}~~~~
    \subfloat{\includegraphics[width = 0.8in,height=1in]{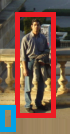}}~~~~
     \subfloat{\includegraphics[width = 0.8in,height=1in]{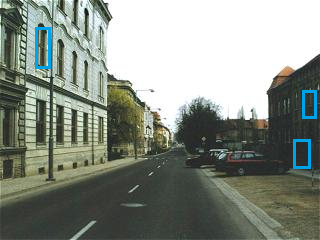}}~~~~
     \subfloat{\includegraphics[width = 0.8in,height=1in]{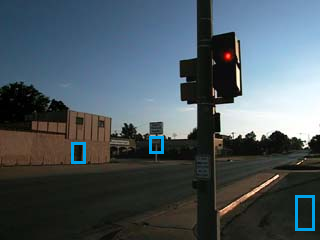}}\\
    \subfloat{(d)}~~~~\subfloat{\includegraphics[width = 0.8in,height=1in]{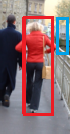}}~~~~
    \subfloat{\includegraphics[width = 0.8in,height=1in]{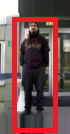}}~~~~
    \subfloat{\includegraphics[width = 0.8in,height=1in]{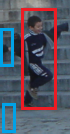}}~~~~
    \subfloat{\includegraphics[width = 0.8in,height=1in]{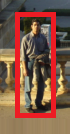}}~~~~
      \subfloat{\includegraphics[width = 0.8in,height=1in]{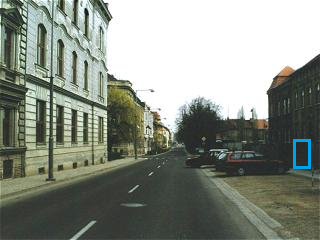}}~~~~
      \subfloat{\includegraphics[width = 0.8in,height=1in]{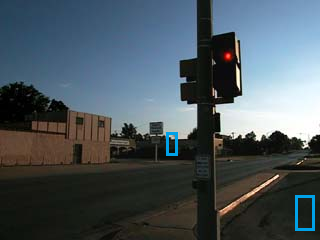}}\\
      \subfloat{(e)}~~~~\subfloat{\includegraphics[width = 0.8in,height=1in]{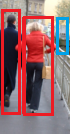}}~~~~
        \subfloat{\includegraphics[width = 0.8in,height=1in]{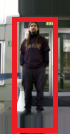}}~~~~
        \subfloat{\includegraphics[width = 0.8in,height=1in]{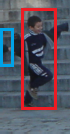}}~~~~
        \subfloat{\includegraphics[width = 0.8in,height=1in]{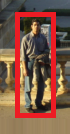}}~~~~
          \subfloat{\includegraphics[width = 0.8in,height=1in]{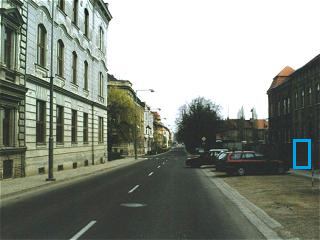}}~~~~
          \subfloat{\includegraphics[width = 0.8in,height=1in]{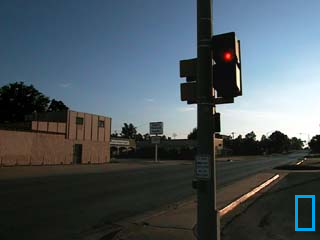}}\\
  \subfloat{(f)}~~~~\subfloat{\includegraphics[width = 0.8in,height=1in]{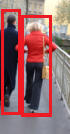}}~~~~
        \subfloat{\includegraphics[width = 0.8in,height=1in]{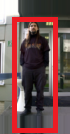}}~~~~
        \subfloat{\includegraphics[width = 0.8in,height=1in]{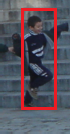}}~~~~
        \subfloat{\includegraphics[width = 0.8in,height=1in]{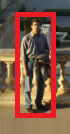}}~~~~
          \subfloat{\includegraphics[width = 0.8in,height=1in]{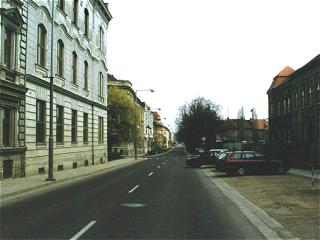}}~~~~
          \subfloat{\includegraphics[width = 0.8in,height=1in]{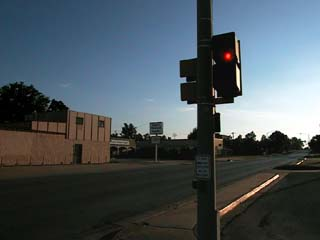}}\\        
     \caption{Pedestrian detection over testing set of INRIA dataset: (a) Pedestrian testing sample images, (b) Bounding box obtained by Pin-SVM (c) Bounding box obtained by Pin-TWSVM and (d) Bounding box obtained by SVMPI, (e) Bounding box obtained by TWSVMPI, and (f) Bounding box obtained by proposed Pin-TWSVMPI where red and blue bounding box represents the positive and negative detection of pedestrian respectively.}
     \label{Fig:pedestrain_bounding_box}
    \end{figure}

    In literature, there are two established methodologies for evaluating pedestrian detection. One of them is per-window performance, and another one is per-image measure. In \cite{dollar2009pedestrian}, the authors have shown that in practice per-window measure could fail to predict per-image performance. Therefore, in this paper, to evaluate the performance of proposed Pin-TWSVMPI for pedestrian detection, we have used the per-image measure method \cite{dollar2009pedestrian}. Generally, a detection system always has ground truth BB ($BB_g$), and it needs to take in an image as input and return a detected Bounding Box ($BB_d$) and a score or confidence for each detection. If the area of a detected BB ($BB_d$) and a ground truth BB ($BB_g$) overlap sufficiently then, they form a potential match. The overlap area between $BB_d$ and $BB_g$ is calculated as  
    \begin{eqnarray}
    ar_o=\frac{area(BB_d\cap BB_g)}{area(BB_d\cup BB_g)}>0.5.
    \end{eqnarray}
    Here, each $BB_d$ and $BB_g$ is matched at most once. The $BB_d$ and $BB_g$ which are not matched to each other are count as false positive and false negative respectively. To analyze different methods, we plot miss rate against false positives per-image(FPPI) in log scale (lower curves indicate better performance) as shown in fig. \ref{Fig:pedestrian_result}. This figure concludes that the privileged information plays a crucial role in pedestrian detection like frameworks where privileged information is already available. Figure \ref{Fig:pedestrain_bounding_box} shows the bounding box obtained by various methods discussed in this paper over testing images. 
  
  \begin{figure}[htp]
   \centering
   \subfloat{(a)}~~~~\subfloat{\includegraphics[width = 0.7in,height=1in]{pos1.jpg}}~~~~
   \subfloat{\includegraphics[width = 0.7in,height=1in]{pos2.jpg}}~~~~
   \subfloat{\includegraphics[width = 0.7in,height=1in]{pos3.jpg}}~~~~
   \subfloat{\includegraphics[width = 0.7in,height=1in]{pos4.jpg}}~~~~
   \subfloat{\includegraphics[width = 0.7in,height=1in]{pos5.jpg}}~~~~
   \subfloat{\includegraphics[width = 0.7in,height=1in]{pos6.jpg}}  \\
  \subfloat{(b)}~~~~\subfloat{\includegraphics[width = 0.7in,height=1in]{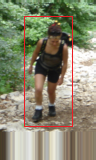}}~~~~
   \subfloat{\includegraphics[width = 0.7in,height=1in]{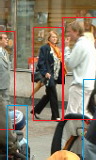}}~~~~
   \subfloat{\includegraphics[width = 0.7in,height=1in]{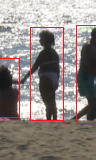}}~~~~
   \subfloat{\includegraphics[width = 0.7in,height=1in]{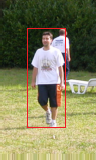}}~~~~
   \subfloat{\includegraphics[width = 0.7in,height=1in]{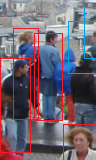}}~~~~
   \subfloat{\includegraphics[width = 0.7in,height=1in]{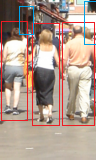}}
    \caption{Pedestrian detection over training set of INRIA dataset with Pin-TWSVMPI: (a) Pedestrian training sample images and (b) Bounding box obtained by proposed Pin-TWSVMPI where red and blue bounding box represents the positive and negative detection of pedestrian respectively.}
   \end{figure}
   
 \subsection{Performance on Handwritten Digit Recognition}
\begin{figure}[htp]
 \centering
 \subfloat{Complete Image}~~~~~~~~\subfloat{Resized Image}\\
 \subfloat{\includegraphics[width = 0.41in,height=0.4in]{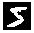}}~~~~~~~~\subfloat{\includegraphics[width = 0.5in,height=0.5in]{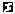}}\\
  \subfloat{\includegraphics[width = 0.41in,height=0.4in]{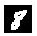}}~~~~~~~~\subfloat{\includegraphics[width = 0.5in,height=0.5in]{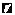}}
  \caption{Sample digit image of first 5 and 8 along with resized image.}
  \label{Fig:MNIST_sample}
 \end{figure}

 MNIST contains 5,522 and 5,652 images of 5 and 8, respectively. The size of images is $28\times 28$ pixel. Classifying these two digits is an easy problem. Therefore, following the experimental setup of \cite{vapnik2009new}, to make it more difficult we resized the digits to $10\times 10$ pixel images. Figure \ref{Fig:MNIST_sample} shows a sample images of original $28\times 28$ image and corresponding $10\times 10$ image. We choose 100 images as a training set, $4000$ images as a validation set (for tuning the parameters) and the rest 1866 images as the test set.
 
\begin{figure}[htp]
  \centering
  \includegraphics[width = 3in,height=1.5in]{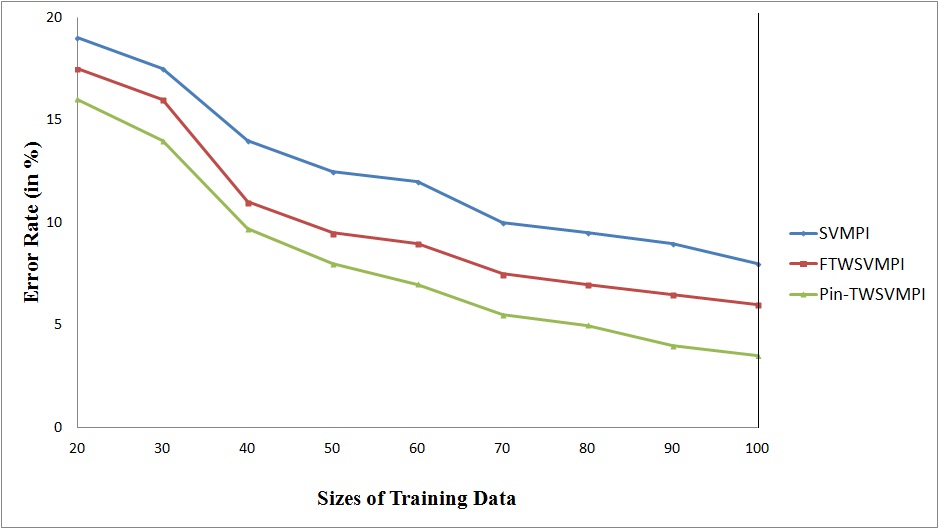}
   \caption{Error rate comparison among SVMPI, FTWSVMPI and Pin-TWSVMPI.}
   \label{Fig:Error_rate_MNIST}
  \end{figure}

\begin{figure}[htp]
    \centering
    \includegraphics[width = 3in,height=1.5in]{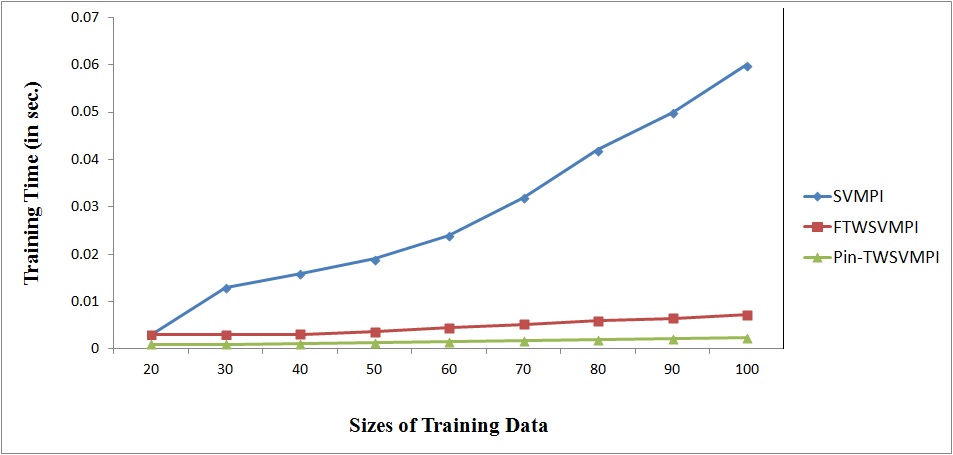}
     \caption{Training time comparison among SVMPI, FTWSVMPI and Pin-TWSVMPI.}
     \label{Fig:train_time_MNIST}
    \end{figure}

 In Fig. \ref{Fig:Error_rate_MNIST}, we present the results by varying the number of training data. From the fig \ref{Fig:Error_rate_MNIST}, we can observe that proposed Pin-TWSVMPI outperforms SVMPI and FTWSVMPI in all cases. The average error rate of Pin-TWSVMPI is $1.92222$ and $4.31111$ lower than that of FTWSVMPI and SVMPI respectively. These results above show that the classifier decided by Pinball loss function is superior to ones by a hinge loss function. In Fig. \ref{Fig:train_time_MNIST}, we show the comparison of training time among the discussed algorithms. More importantly, from fig. \ref{Fig:train_time_MNIST}, we can find the training time speed of Pin-TWSVMPI is much faster than STWSVMPI and SVMPI.
 
\section{Conclusions}
\label{conclusion}
In this paper, we have proposed a fast Twin Support Vector Machine based on privileged information with pinball loss classifier (termed as Pin-TWSVMPI) and have shown their applications over pedestrian detection and handwritten digit recognition. Some major properties of pinball loss, i.e., noise insensitivity and re-sampling ability over other loss functions motivate us to build Pin-TWSVM using privileged information. As far as we know, it is the first time that privileged information has been incorporated to improve the performance of Pin-TWSVM classifier. The privileged information is identified from different individual expert's knowledge, which enhanced the generalization performance of the proposed framework. In order to incorporate the SRM principle along with ERM principle, we introduced a regularization term in the objective function of Pin-TWSVMPI. We have also used SMO to solve the dual problem of proposed formulation which helps in solving Pin-TWSVMPI efficiently. This framework also provides the freedom to integrate information obtained from multiple expert's knowledge without building a separate new model which also improves the training time of the classifier. We have also implemented the novel method given in \cite{pal2018learning} which extracts privileged information by avoiding to solve two additional QPPs as in FTWSVMPI where expert's knowledge is not available. The experiment results on several UCI benchmark datasets show that our proposed method achieves better classification accuracy to that of other algorithms, i.e., Pin-TWSVM and TWSVMPI with considerably lesser computational time. As the applications to the proposed Pin-TWSVMPI model, we have also performed experiments for pedestrian detection over INRIA dataset and handwritten digit recognition over MNIST dataset. As a line of future research, it would be interesting to see how shallow learning algorithms such as XGBoost would improve the performance of the proposed model. 

\textbf{Conflict of interest}\\
There are no conflicts of interest in this study.\\

%

%

\section*{Conflict of Interest}
We authors hereby declare that we do not have any conflict of interest with the content of this manuscript.

\section*{Funding}
Second author gratefully acknowledges the financial support received from the South Asian University, India in the form of scholarship. 

\section*{Ethical Approval}
This article does not contain any studies with human participants or animals performed by any of the authors.

\end{document}